# Guider l'attention dans les modèles de séquence à séquence pour la prédiction des actes de dialogue

Emile Chapuis*
emile.chapuis@telecom-paris.fr
Telecom Paris

Pierre Colombo*
pierre.colombo@telecom-paris.fr
pierre.colombo@ibm.com
IBM/Telecom Paris

Matteo Manica
IBM Research Zurich
tte@zurich.ibm.com

Giovanna Varni
Telecom Paris
giovanna.varni@telecom-paris.fr

Emmanuel Vignon
IBM France
emmanuel.vignon@fr.ibm.com

Chloé Clavel
Telecom Paris
chloe.clavel@telecom-paris.fr

Translated from paper Guiding attention in Sequence-to-sequence models for Dialogue Act prediction.

## ABSTRACT

La prédiction d'actes de dialogue (AD) basés sur le dialogue conversationnel est un élément clé dans le développement des agents conversationnels. La prédiction précise des AD nécessite une modélisation précise à la fois de la conversation et des dépendances globales des AD. Nous utilisons les approches de séquence à séquence (seq2seq) largement adoptées dans la traduction automatique neurale (NMT) pour améliorer la modélisation de la séquentialité des AD. Les modèles seq2seq sont connus pour apprendre les dépendances globales complexes alors que les approches actuellement proposées utilisant des champs aléatoires conditionnels linéaires (CRF) ne modélisent que les dépendances locales des AD. Dans ce travail, nous introduisons un modèle seq2seq adapté à la classification AD en utilisant : un codeur hiérarchique, un nouveau mécanisme *attention guidée* et la recherche de faisceau appliquée à la fois à l'apprentissage et à l'inférence. Par rapport à l'état de l'art, notre modèle ne nécessite pas de caractéristiques artisanales et est formé de bout en bout. En outre, l'approche proposée obtient un score de précision inégalé de 85% pour la SwDA et un score de précision de pointe de 91,6% pour la MRDA.

## 1 INTRODUCTION

Dans la recherche sur le traitement du langage naturel, le concept d'acte de dialogue (AD) joue un rôle important. Les AD sont des étiquettes sémantiques associées à chaque énoncé dans un dialogue conversationnel qui indiquent l'intention de l'orateur, par exemple, une question, un *backchannel*, une déclaration de non-opinion, une opinion de déclaration. Une des clés du dialogue modèle consiste à détecter l'intention de l'orateur : identifier correctement une question donne un indice important pour produire une réponse appropriée. Comme on peut l'observer dans le tableau 1, la classifi-cation AD repose sur son aspect conversationnel, c'est-à-dire que la prédiction de la AD d'un énoncé nécessite la connaissance des phrases précédentes et des étiquettes d'actes qui leur sont associées. Par exemple, si un orateur pose une question, son interlocuteur répondra par une réponse, de même, un "salut" ou un "au revoir" sera suivi d'un acte de dialogue similaire. Cela signifie que dans une conversation, il y a une structure séquentielle dans les actes de dialogue émis. Ceci pose la base pour l'adoption d'une nouvelle perspective sur le problème de la classification des AD, c'est-à-dire d'une tâche de multi-classification à une tâche d'étiquetage séquentiel.

| Speaker | Utterance |
|---------|-----------|
| A | Is there anyone who doesn't know Nancy? |
| A | Do you - Do you know Nancy ? |
| B | Me? |
| B | Mm-hmm |
| B | I know Nancy |

Table 1: Exemple de conversation tirée de Switchboard. A parle avec B.

**Limitations des modèles actuels :** Les modèles actuels de pointe reposent sur l'utilisation d'un champ aléatoire conditionnel (CRF) linéaire combiné à un codeur récurrent basé sur un réseau neuronal [4, 15, 21] pour modéliser les dépendances séquentielles de l'AD. Malheureusement, ces approches ne prennent en compte que les dépendances locales entre deux actes de dialogue adjacents. Par exemple, si nous considérons l'exemple du tableau 1, nous pouvons voir que la dernière déclaration "Je connais Nancy" est une réponse à la première question "Y a-t-il quelqu'un qui ne connaît pas Nancy" et que la connaissance du *backchannel* précédent ne facilite pas la prédiction du dernier acte de dialogue. Par conséquent, nous devons tenir compte des dépendances entre les étiquettes dont la portée est plus large que deux énoncés successifs. Dans la traduction automatique neuronale (NMT), le problème des dépendances globales a été abordé à l'aide des modèles seq2seq [27] qui suivent le cadre de l'encodeur-décodeur. L'encodeur intègre une phrase d'entrée dans un vecteur caché unique qui contient à la fois des dépendances globales et locales, et le vecteur caché est ensuite décodé pour produire une séquence de sortie. Dans ce travail, nous proposons une architecture seq2seq adaptée à la classification AD, ouvrant la voie à d'autres innovations inspirées par les progrès de la recherche sur les NMT.

---

*Les deux auteurs ont contribué à part égale à cette recherche.

**Contributions:** Dans ce travail, (1) nous formalisons le problème de la prédiction de l'acte de dialogue d'une nière qui souligne les relations entre la classification AD et la NMT, (2) nous démontrons que l'architecture seq2seq convient mieux à la tâche de classification AD et (3) nous présentons un modèle seq2seq exploitant les techniques NMT qui atteint une précision de 85%, surpassant l'état de l'art par une marge d'environ 2%, sur le Corpus de la loi sur le dialogue avec les standards téléphoniques (SwDA) [25] et un score de précision de 91,6% sur la loi sur le dialogue avec les enregistreurs de réunion (MRDA). Ce modèle seq2seq exploite un codeur hiérarchique doté d'un nouveau mécanisme *d'attention guidée* qui s'adapte à notre environnement sans aucune caractéristique artisanale. Nous affinons notre seq2seq en utilisant un objectif d'entraînement au niveau de la séquence faisant appel à l'algorithme de recherche de faisceau. À notre connaissance, c'est l'un des premiers modèles de seq2seq proposés pour la classification AD.

## 2 CONTEXTE
### 2.1 Classification de AD

Plusieurs approches ont été proposées pour résoudre le problème de la classification des AD. Ces méthodes peuvent être divisées en deux catégories différentes. La première catégorie de méthodes repose sur la classification indépendante de chaque énoncé à l'aide de diverses techniques, telles que les HMM [24], les SVM [26] et les réseaux bayésiens [9]. La deuxième classe, qui permet d'obtenir de meilleures performances, exploite le contexte pour améliorer les performances du classificateur en utilisant des approches d'apprentissage approfondi pour saisir les dépendances contextuelles entre les phrases d'entrée [3, 10]. Un autre perfectionnement de la classification basée sur le contexte d'entrée est la modélisation des dépendances entre les phrases d'entrée. Cette tâche est abordée comme une classification basée sur des séquences où les AD de sortie sont considérées comme une séquence d'AD [4, 12, 15? ].

Deux critères de référence classiques sont adoptés pour évaluer les systèmes de classification des AD : le "Switchboard Dialogue Act Corpus" (SwDA) [25] et le "Meeting Recorder Dialogue Act" (MRDA) [7]. Les techniques de pointe permettent d'atteindre une précision de 82,9% [15]. Pour saisir les dépendances contextuelles en entrée, elles adoptent un encodeur hiérarchique et un CRF pour modéliser les dépendances entre les marques. La principale limite de l'architecture susmentionnée est qu'un modèle CRF linéaire ne peut capturer que les dépendances au niveau local et ne peut pas capturer les dépendances non locales. Dans cet article, nous abordons cette question avec un seq2seq en utilisant un mécanisme d'attention guidée.

### 2.2 Modèles Seq2seq

Les modèles Seq2seq ont été appliqués avec succès aux NMT, où la modélisation des dépendances non locales est un défi crucial. La classification des AD peut être considérée comme un problème lorsque l'objectif est de faire correspondre une séquence d'énoncés à une séquence d'AD. Ainsi, elle peut être formulée comme un problème de séquence à séquence très similaire à celui des NMT. L'architecture générale de nos modèles seq2seq [27] suit une approche classique d'encodeur-décodeur avec attention [18]. Nous utilisons des cellules GRU [5], car elles sont plus rapides à entrainer que celles des LSTM [8]. Des progrès récents ont amélioré à la fois le processus d'apprentissage et le processus d'inférence, produisant des séquences plus cohérentes grâce à des pertes de niveau de séquence Wiseman and Rush [28] et à divers paramètres de recherche par faisceau [29]. Le paramètre le plus proche où les modèles seq2seq ont été utilisés avec succès est l'analyse des dépendances [16], où les dépendances de sortie sont cruciales pour obtenir des performances de pointe. Dans notre travail, nous adaptons les techniques NMT aux spécificités de la classification AD.

## 3 PROBLÉMATIQUE
### 3.1 La classification des AD comme un problème de NMT

Tout d'abord, définissons les notations mathématiques que nous adopterons dans ce travail. Nous avons un ensemble $D$ de conversations, c'est-à-dire $D = (C_1, C_2, \ldots, C_{|D|})$ avec $Y = (Y_1, Y_2, \ldots, Y_{|D|})$ l'ensemble correspondant d'étiquettes AD. Une conversation $C_i$ est une séquence d'énoncés, à savoir $C_i = (u_1, u_2, \ldots, u_{|C_i|})$ avec $Y_i = (y_1, y_2, \ldots, y_{|C_i|})$ la séquence correspondante d'étiquettes AD. Ainsi, chaque énoncé $u_i$ est associé à une étiquette AD unique $y_i \in \mathcal{Y}$ où $\mathcal{Y}$ est l'ensemble de tous les actes de dialogue possibles. Enfin, un énoncé $u_i$ peut être considéré comme une séquence de mots, c'est-à-dire $u_i = (\omega_1^i, \omega_2^i, \ldots, \omega_{|u_i|}^i)$. En NMT, le but est d'associer pour toute phrase $X^{l_1} = (x_1^{l_1}, \ldots, x_{|X^{l_1}|}^{l_1})$ dans la langue $l_1$ une phrase $X^{l_2} = (x_1^{l_2}, \ldots, x_{|X^{l_2}|}^{l_2})$ dans la langue $l_2$ où $x_i^{l_k}$ est le mot $i$ dans la phrase dans la langue $l_k$. En utilisant ce formalisme, il est aisé de constater deux grandes similitudes ($\mathbb{S}_1, \mathbb{S}_2$) entre la classification AD et la NMT. ($\mathbb{S}_1$) Dans la classification NMT et AD, le but est de maximiser la probabilité de la séquence de sortie étant donné la séquence d'entrée ($P(X^{l_2}|X^{l_1})$ contre $P(Y_i|C_i)$). ($\mathbb{S}_2$) Pour les deux tâches, il existe de fortes dépendances entre les unités composant à la fois les séquences d'entrée et de sortie. Dans la classification NMT, ces unités sont des mots ($x_i$ et $y_i$), dans la classification AD, ces unités sont des énoncés et des étiquettes AD ($u_i$ et $y_i$).

### 3.2 Particularités de la classification AD

Si les classifications NMT et AD sont similaires à certains égards, trois différences apparaissent immédiatement ($\mathbb{D}_i$). ($\mathbb{D}_1$) Dans la classification NMT, les unités d'entrée $x_i$ représentent des mots, dans la classification AD $u_i$ sont des séquences d'entrée composées de mots. Considérer l'ensemble de toutes les séquences possibles comme entrée (la prise en compte du contexte conduit à une performance supérieure) implique que la dimension de l'espace d'entrée soit plusieurs ordres de grandeur plus grande que celle d'un NMT standard. ($\mathbb{D}_2$) En AD, nous avons un alignement parfait entre les séquences d'entrée et de sortie (d'où $T = T\prime$). Certaines langues, par exemple le français, l'anglais et l'italien, ont un alignement partiel, mais dans la classification AD, nous avons un fort mapping entre $y_i$ et $x_i$. ($\mathbb{D}_3$) En NMT, l'espace d'entrée (nombre de mots dans $l_1$) est approximativement de la même taille que l'espace de sortie (nombre de mots dans $l_2$). Dans notre cas, l'espace de sortie (nombre d'AD $|\mathcal{Y}| < 100$) a une taille limitée, avec une dimension qui est de plusieurs ordres de grandeur inférieure à celle de l'espace d'entrée.

Dans ce qui suit, nous proposons une architecture seq2seq pour la classification AD qui exploite ($\mathbb{D}_1$) en utilisant un encodeur hiérarchique, ($\mathbb{D}_2$) par un mécanisme d'attention guidée et ($\mathbb{D}_3$) en utilisant la recherche de faisceau pendant l'entraînement et l'inférence, en tirant parti de la dimension limitée de l'espace de sortie.

## 4 MODELS

Dans Seq2seq, le codeur prend une séquence de phrases et la représente comme un vecteur unique $H_i \in \mathcal{R}^d$ et la transmet ensuite au décodeur pour les générations des AD.

### 4.1 Encodeurs

Dans cette section, nous présentons les différents codeurs que nous considérons dans nos expériences. Nous exploitons la structure hiérarchique du dialogue pour réduire la taille de l'espace d'entrée ($\mathbb{D}_1$) et pour préserver la structure des mots et des phrases. Pendant la formation et l'inférence, la taille du contexte est fixée à $T$. Formellement, un codeur prend en entrée un nombre fixe d'énoncés ($u_{i-T}, .., u_i$) et produit un vecteur $H_i \in \mathcal{R}^d$ qui servira à initialiser l'état caché du décodeur. Le premier niveau de l'encodeur calcule $\mathcal{E}_{u_t}$, un encastrement de $u_t$ basé sur les mots composant l'énoncé, et les niveaux suivants calculent $H_i$ sur la base de $\mathcal{E}_{u_t}$.

**Encodeur RNN:** L'encodeur RNN (VGRU$_E$) introduit par Sutskever et al. [27] est considéré comme un encodeur de base. Dans l'encodeur vanille $\mathcal{E}_{u_i} = \frac{1}{|u_i|} \sum_{k=1}^{|u_i|} \mathcal{E}_{w_k^i}$ où $\mathcal{E}_{w_k^i}$ est un encastrement de $w_k^i$. Pour mieux modéliser les dépendances entre des énoncés consécutifs, nous utilisons un GRU bidirectionnel [5] :

$$\begin{aligned}
\overrightarrow{h_{i-T}^s} &= \overleftarrow{h_{i-T}^s} = \overrightarrow{0} \\
\overrightarrow{h_t^s} &= \overrightarrow{GRU}(\mathcal{E}_{u_t}), t \in [i-T, i] \\
\overleftarrow{h_t^s} &= \overleftarrow{GRU}(\mathcal{E}_{u_t}), t \in [i, i-T] \\
H_i &= [\overleftarrow{h_i^s}, \overrightarrow{h_i^s}]
\end{aligned} \quad (1)$$

**Encodeurs hiérarchiques:** L'encodeur vanille peut être amélioré en calculant $\mathcal{E}_{u_i}$ en utilisant bi-GRU. Cet encodeur hiérarchique (HGRU) est conforme à celui introduit par Sordoni et al. [23]. Formellement, $\mathcal{E}_{u_i}$ est défini comme suit :

$$\begin{aligned}
\overrightarrow{h_0^w} &= \overleftarrow{h_0^w} = \overrightarrow{0} \\
\overrightarrow{h_t^w} &= \overrightarrow{GRU}(\mathcal{E}_{w_t^i}), t \in [1, |u_i|] \\
\overleftarrow{h_t^w} &= \overleftarrow{GRU}(\mathcal{E}_{w_t^i}), t \in [|u_i|, 1] \\
\mathcal{E}_{u_i} &= [\overleftarrow{h_{|u_i|}^w}, \overrightarrow{h_{|u_i|}^w}]
\end{aligned} \quad (2)$$

$H_i$ est ensuite calculé à l'aide de l'équation 1. Intuitivement, la première couche du GRU (Equation 2) modélise les dépendances entre les mots (l'état caché du GRU au niveau du mot est réinitialisé à chaque nouvelle prononciation), et la deuxième couche modélise les dépendances entre les prononciations.

**Encodeur Personnel Hiérarchique:** Dans SwDA, un tour de parole peut être divisé en plusieurs énoncés. Par exemple, si le locuteur A interagit avec le locuteur B, nous pouvons rencontrer la séquence (AAABBBAA). Nous proposons un nouvel encodeur personnel hiérarchique (PersoHGRU) pour mieux modéliser les dépendances entre le locuteur et l'extinction. Nous introduisons une couche de personnalité entre les niveaux du mot et de la phrase, voir Figure 1 :

$$\begin{aligned}
\overrightarrow{h_t^p} &= \begin{cases} \overrightarrow{0} & \text{if } t \text{ and } t-1 \text{ have different speakers} \\ \overrightarrow{GRU}(\mathcal{E}_{u_{t-1}}) \end{cases} \\
\overleftarrow{h_t^p} &= \begin{cases} \overrightarrow{0} & \text{if } t \text{ and } t+1 \text{ have different speakers} \\ \overleftarrow{GRU}(\mathcal{E}_{u_{t+1}}) \end{cases} \\
\mathcal{E}_{u_k}^p &= [\overrightarrow{h_k^p}, \overleftarrow{h_k^p}] \quad \forall k \in [i-T, i]
\end{aligned} \quad (3)$$

On obtient alors $H_i$ en suivant l'équation 1 où $\mathcal{E}_{u_i}$ est remplacé par $\mathcal{E}_{u_i}^p$.

### 4.2 Décodeurs

Dans cette section, nous présentons les différents décodeurs que nous comparons dans nos expériences. Nous présentons une nouvelle forme d'attention que nous appelons *attention guidée*. *Attention guidée* exploite l'alignement parfait entre les séquences d'entrée et de sortie ($\mathbb{D}_2$). Le décodeur calcule la probabilité de la séquence d'étiquettes de sortie sur la base de :

$$p(y_{i-T}, \ldots, y_i | u_{i-T}, \ldots, u_i) = \prod_{k=i-T}^{i} p(y_k | H_i, y_{k-1}, \ldots, y_{i-T}) \quad (4)$$

see Equation 1.

**Décodeur:** Le décodeur (VGRU$_D$) est similaire à celui introduit par Sutskever et al. [27].

**Décodeurs avec attention:** En NMT, le mécanisme d'attention force le modèle seq2seq à apprendre à se concentrer sur des parties spécifiques de la séquence à chaque fois qu'un nouveau mot est généré et à laisser le décodeur aligner correctement la séquence d'entrée avec la séquence de sortie. Dans notre cas, nous suivons l'approche décrite par Bahdanau et al. [1] et nous définissons le vecteur de contexte comme:

$$c_k = \sum_{j=i-T}^{i} \alpha_{j,k} h_j^s \quad (5)$$

où $\alpha_{j,k}$ indique la correspondance entre les entrées autour de la position $k$ et les sorties à la position $j$. Comme nous avons un alignement parfait ($\mathbb{D}_2$), nous savons a priori sur quelle séquence le décodeur doit se concentrer davantage à chaque pas de temps. En tenant compte de cet aspect du problème, nous proposons trois mécanismes d'attention différents.

**Attention:** Cette attention représente notre mécanisme d'attention de base et c'est celui proposé par Bahdanau et al. [1], où :

$$\alpha_{j,k} = softmax(a(h_{k-1}^{Dec}, h_j^s)) \quad (6)$$

and $a$ is parametrized as a feedforward neural network.

**Attention fortement guidée:** Le *attention fortement guidée* force le décodeur à se concentrer uniquement sur les $u_i$ tout en prédisant les $y_i$:

$$\alpha_{j,k} = \begin{cases} 0, & \text{if } k \neq j \\ 1, & \text{otherwise} \end{cases} \quad (7)$$

**attention faiblement guidée:** L' *attention faiblement guidée* guide le décodeur pour qu'il se concentre principalement sur le $u_i$ tout en prédisant le $y_i$, mais lui permet de se concentrer de manière

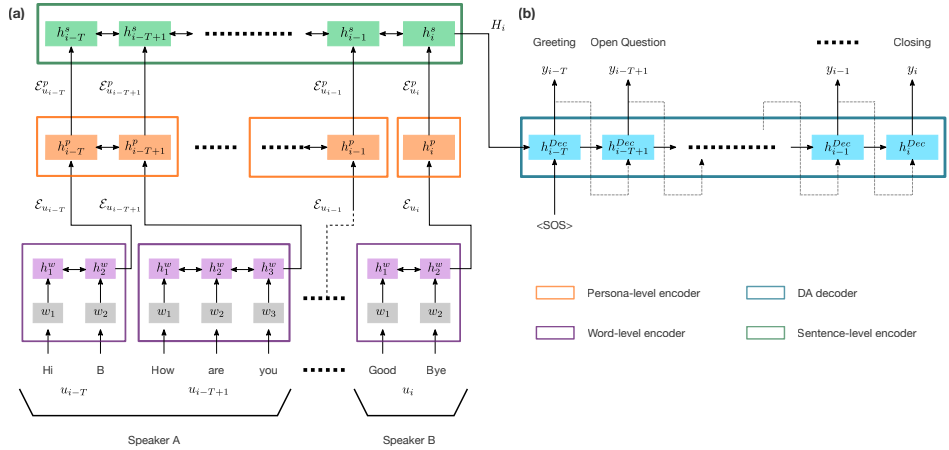

Figure 1: Architecture modèle Seq2seq pour la classification AD. (a) L'encodeur est composé de trois niveaux différents représentant un niveau hiérarchique différent dans le dialogue. Les énoncés sont encodés au niveau du mot (violet), du personnage (orange) et de la phrase (vert). b) Le décodeur (bleu) est chargé de générer pour chaque énoncé un AD exploitant le dernier état du codeur comme état initial caché.

limitée sur d'autres parties de la séquence d'entrée.

$$\widetilde{\alpha}_{j,k} = \begin{cases} a(h_{k-1}^{Dec}, h_j^s), & \text{if } k \neq j \\ 1 + a(h_{k-1}^{Dec}, h_j^s), & \text{otherwise} \end{cases} \quad (8)$$

$$\alpha_{j,k} = softmax(\widetilde{\alpha}_{j,k}) \quad (9)$$

où $a$ est paramétré comme un réseau neuronal de feedforward.

### 4.3 Entrainement et inférence

Dans cette section, nous décrivons la formation et les stratégies d'inférence utilisées pour nos modèles. Un modèle seq2seq vise à trouver la meilleure phrase pour une phrase source donnée. Cela pose un problème de calcul lorsque la taille du vocabulaire de sortie est importante, car même en utilisant la recherche par faisceaux, il est coûteux d'explorer plusieurs chemins. Comme la taille de notre vocabulaire de sortie est limitée ($\mathbb{D}_3$), nous ne sommes pas confrontés à ce problème et nous pouvons utiliser la recherche par faisceau à la fois pendant l'entraînement et l'inférence.

**Recherche par faisceaux :** Dans notre travail, nous mesurons la probabilité de séquence sur la base de la formule suivante :

$$s(\tilde{y}^k, u_i) = \frac{\log P(\tilde{y}^k | u_i)}{lp(\tilde{y}^k)} \quad (10)$$

où $u_i = (u_{i-T}, \ldots, u_i)$ et $\tilde{y}^k = (\tilde{y}_{i-T}, \ldots, \tilde{y}_{i-T+k})$ est la cible actuelle, et $lp(\tilde{y}) = \frac{(5+|\tilde{y}|)^\alpha}{(5+1)^\alpha}$ est le coefficient de normalisation de la longueur [29]. À chaque pas de temps, les séquences les plus probables $B$ sont conservées ($B$ correspondant à la taille du faisceau).

**Objectif d'entraînement:** Pour la formation, nous suivons Wiseman and Rush [28] et nous entraînons notre modèle jusqu'à la convergence avec une perte de niveau symbolique et nous l'affinons en minimisant le risque attendu $\mathcal{L}_{\text{RISK}}$ défini comme:

$$\mathcal{L}_{\text{RISK}} = \sum_{\tilde{y} \in U(C_i)} \frac{cost(\tilde{y}, y_i) p(\tilde{y}|u_i)}{\sum_{\tilde{y}' \in U(u_i)} p(\tilde{y}|u_i)} \quad (11)$$

où $U(u_i)$ est l'ensemble des séquences générées par le modèle en utilisant un algorithme de recherche par faisceaux pour l'entrée $u_i$, et $cost(\tilde{y}, y_i)$ est défini, pour une séquence candidate donnée $\tilde{y}$ et une cible $y_i$, comme:

$$cost(\tilde{y}, y_i) = \begin{cases} 1 & \text{if } \tilde{y}_i = y_i \\ 0 & \text{otherwise} \end{cases} \quad (12)$$

### 4.4 GRU/HGRU CRF baseline

Les modèles de pointe utilisent des champs aléatoires conditionnels qui modélisent les dépendances entre les AD situées au-dessus d'un encodeur GRU ou HGRU qui a calculé un encastrement d'un nombre variable de phrases prononcées. Nous avons implémenté notre propre (Baseline$_{\text{CRF}}$) suivant le travail de Kumar et al. [12]:

$$p(y_i, \ldots, y_{i-T}, u_i, \ldots, u_{i-T}; \theta) = \frac{\prod_{t=i}^{T} \psi(y_{t-1}, y_t, \phi(o_t); \theta)}{\sum_{\mathcal{Y}} \prod_{t=i}^{T} \psi(y_t, y_{t-1}, \phi(o_t); \theta)} \quad (13)$$

Ici, $\theta$ est l'ensemble des paramètres correspondant à la couche CRF, et $\psi$ est la fonction de fonctionnalité, nous fournissant des potentiels unaire et par paire. Soit $\phi \colon \mathbb{R}^H \to \mathbb{R}^{|\mathcal{Y}|}$ la représentation dense de la sortie de chaque énoncé fournie par le codeur. $\phi$ peut être considéré comme la fonction de caractéristique unaire.

## 5 PROTOCOLE EXPÉRIMENTAL

Dans cette section, nous décrivons les protocoles expérimentaux adoptés pour l'évaluation de notre approche.

### 5.1 Jeux de données

Nous considérons deux ensembles de données classiques pour la classification des actes de dialogue : Le Corpus de l'acte de dialogue sur le standard téléphonique et le MRDA. Comme nos modèles génèrent explicitement une séquence de AD, nous calculons la précision du dernier AD généré. Les deux ensembles de données sont déjà segmentés en paroles et chaque parole est segmentée en mots. Pour chaque ensemble de données, nous divisons chaque conversation $C_i$ en une séquence d'énoncés de longueur $T = 5^1$.

**SwDA:** Le corpus Switchboard-1 est un corpus de parole téléphonique [25], composé d'environ 2.400 conversations téléphoniques bilatérales entre 543 locuteurs avec environ 70 sujets de conversation fournis. L'ensemble de données comprend des informations sur les locuteurs et les sujets et possède 42 AD différents. Dans cet ensemble de données, la dépendance globale joue un rôle clé en raison de la grande quantité de canaux de retour (19 %), d'abandons ou de sorties (5 %) et d'actes non interprétables (1 %). Dans ce contexte, tout modèle qui ne prend en compte que les dépendances locales ne parviendra pas à extraire les informations permettant de distinguer les AD ambiguës. Pour la matrice de confusion, nous suivons Li et al. [15] et la présentons pour 10 AD seulement : déclaration-non-opinion (sd), *backchannel* (b), déclaration-opinion (sv), fermeture conventionnelle (fc), wh-question (qw), accusé de réception de la réponse (bk), couverture (h), question ouverte (qo), autres réponses (no), remerciements (ft).

**MRDA:** MRDA : Le corpus ICSI Meeting Recorder Dialogue Act [22] contient 72 heures de réunions multipartites qui ont d'abord été converties en conversations de 75 mots, puis annotées à la main avec les AD à l'aide du jeu des AD du Meeting Recorder Dialogue Act. Dans ce travail, nous utilisons 5 AD, c'est-à-dire des déclarations (s), des questions (q), des saisies de sol (f), *backchannel* (b), disruption (d).

**Entrainement/Développement/Test séprations** : Pour le SwDA et le MRDA, nous suivons le fractionnement officiel introduit par Stolcke et al. [24]. Ainsi, notre modèle peut être directement comparé à Chen et al. [4], Kumar et al. [12], Li et al. [15], Raheja and Tetreault [21].

### 5.2 Détails d'entrainement

Tous les hyperparamètres ont été optimisés sur l'ensemble de validation en utilisant la précision calculée sur le dernier tag de la séquence. La couche d'incorporation est initialisée avec des vecteurs de mots FastText préformés de taille 300 [2]$^2$ , formés avec des informations sur les sous-mots (sur Wikipédia 2017, le corpus de la base web de l'UMBC et l'ensemble des données de nouvelles de statmt.org), et mis à jour pendant la formation. La sélection des hyperparamètres a été effectuée à l'aide d'une recherche aléatoire sur une grille fixe. Les modèles ont été implémentés dans PyTorch et formés sur un seul NVIDIA P100.

**Paramètres pour SwDA:** Nous avons utilisé Adam optimizer [11] avec un taux d'apprentissage de 0,01, qui est mis à jour à l'aide d'un programmateur avec une patience de 20 époques et un taux de diminution de 0,5. La norme de gradient est réduite à 5,0, la décroissance du poids est fixée à 1e-5, et la décroissance [13] est fixée à 0,2. La longueur maximale de la séquence est fixée à 20. Le modèle le plus performant est un codeur de taille 128 et un décodeur de taille 48. Pour VGRU$_E$, nous utilisons deux couches pour la couche BiGRU. Pour les modèles hiérarchiques, nous utilisons BiGRU avec une seule couche.

**Paramètres pour MRDA:** Nous avons utilisé l'optimiseur AdamW [17] avec un taux d'apprentissage de 0,001, qui est mis à jour à l'aide d'un programmateur avec une patience de 15 époques et un taux de diminution de 0,5. La norme de gradient est réduite à 5,0, la décroissance du poids est fixée à 5e-5, et la décroissance [13] est fixée à 0,3. La longueur maximale de la séquence est fixée à 30. Le modèle le plus performant est un codeur de taille 40 et un décodeur de taille 400. Pour VGRU$_E$ nous utilisons deux couches pour la couche BiGRU, pour les modèles hiérarchiques nous utilisons BiGRU avec une seule couche.

## 6 EXPÉRIENCES & RÉSULTATS

Dans cette section, nous proposons une série d'expériences afin d'étudier les performances de notre modèle par rapport aux approches existantes en ce qui concerne les difficultés soulignées dans l'introduction.

### 6.1 Expérience 1 : les Seq2seq sont-ils mieux adaptés à la prédiction des AD que les CRF ?

L'état de l'art actuel est basé sur les modèles CRF. Dans cette première section, nous visons à comparer un seq2seq avec un modèle basé sur le CRF. Afin de fournir une comparaison équitable, nous effectuons la même recherche de grille pour tous les modèles sur une grille fixe. À cette étape, nous n'utilisons pas l'attention ni la recherche par faisceaux pendant l'entraînement ou l'inférence. Comme le montre le tableau 2, avec un encodeur RNN, le seq2seq surpasse significativement le CRF sur SwDa et MRDA. Avec un HGRU, le seq2seq présente des résultats significativement plus élevés sur SwDA et atteint des performances comparables sur MRDA. Ce comportement suggère qu'un modèle basé sur une architecture seq2seq tend à obtenir un score plus élevé sur la classification AD qu'un modèle basé sur le CRF.

| Models | SwDa | MRDA |
|---|---|---|
| Baseline$_{CRF}$ (+GRU) | 77.7 | 88.3 |
| seq2seq (+GRU) | **81.9** | **88.5** |
| Baseline$_{CRF}$ (+HGRU) | 81.6 | 90.0 |
| seq2seq (+HGRU) | **82.4** | 90.0 |

**Table 2: Précision d'une seq2seq sur l'ensemble dev et Baseline$_{CRF}$ sur SwDA et MRDA. Les résultats en gras montrent des différences significatives (p-value $< 0,01$) selon le test Wilcoxon Mann Whitney effectué sur 10 passages avec différentes graines.**

**Analyse des dépendances globales:** Dans le tableau 3, nous présentons deux exemples où notre seq2seq utilise des informations contextuelles pour désambiguïser l'étiquette et pour prédire l'étiquette correcte. Dans le premier exemple, "It can be a pain" sans contexte peut être interprété à la fois comme une déclaration de non-opinion (sd) ou une déclaration d'opinion (sv). Notre seq2seq utilise le contexte environnant (deux phrases avant) pour désambiguïser et attribuer le label sv . Dans le deuxième exemple, l'étiquette correcte attribuée à "Oh, ok" est un accusé de réception de réponse (bk) et non un canal de retour (b). La principale différence entre bk et b est qu'un énoncé étiqueté bk doit être produit dans un contexte de question-réponse, alors que b est un ***continuiteur***. Dans notre exemple, le contexte global est une situation de question/réponse : le premier intervenant pose une question ("Quelle école est-ce ?"), le second répond ensuite, le premier intervenant répond à la réponse. Cette observation reflète le fait que les modèles CRF ne traitent que les dépendances locales, alors que les modèles seq2seq considèrent également les dépendances globales.

---
[1]$T$ est un hyperparamètre, des expériences ont montré que 5 conduit aux meilleurs résultats.
[2]Dans notre travail, nous nous appuyons sur le même vecteur de mots d'incorporation préformés word2vect [19] au lieu de GloVe [20].

| Utterances | G. | seq2seq | CRF |
|---|---|---|---|
| How long does that take you to get to work? | qw | qw | qw |
| Uh, about forty-five, fifty minutes. | sd | sd | sd |
| How does that work, work out with, uh, storing your bike and showering and all that? | qw | qw | qw |
| Yeah , | b | b | b |
| It can be a pain . | sd | sd | sv |
| It's, it's nice riding to school because it's all along a canal path, uh, | sd | sd | sd |
| Because it's just, it's along the Erie Canal up here. | sd | sd | sd |
| So, what school is it? | qw | qw | qw |
| Uh, University of Rochester. | sd | sd | sd |
| Oh, okay. | bk | bk | b |

**Table 3: Exemple de séquence prédite d'étiquettes tirées de SwDA. seq2seq est notre modèle le plus performant, CRF signifie Baseline$_{CRF}$, G. est le label.**

| Enc. \ Dec. | SwDA | | | |
|---|---|---|---|---|
| | VGRU$_D$ | att. | faib guid. | fort guid. |
| Taille du faisceau | 1 | 1 | 1 | 1 |
| VGRU$_E$ | 81.6 | 82.1 | 82.8 | 82.9 |
| HGRU | 82.4 | 82.3 | 83.1 | **84.0** |
| PersoHGRU | 49.8 | 79.4 | 84.0 | 83.5 |
| | MRDA | | | |
| Enc. \ Dec. | VGRU$_D$ | att. | faib guid. | fort guid. |
| Taille du faisceau | 1 | 1 | 1 | 1 |
| VGRU$_E$ | 88.5 | 88.5 | 88.5 | 88.5 |
| HGRU | 90.0 | 89.9 | 90.0 | **90.2** |
| PersoHGRU | 66.2 | 87.7 | 88.2 | 86.9 |

**Table 4: Précision sur le jeu de développement des différentes combinaisons de codeurs/décodeurs MRDA et SwDA. Pour SwDA, le test de Wilcoxon (10 essais avec différentes graines) a été effectué pour un encodeur HGRU avec un décodeur avec *attention fortement guidée* contre un encodeur HGRU avec *attention faiblement guidée, attention faiblement guidée,* avec attention, sans attention Les tests par paires présentent p-value < 0.01.**

## 6.2 Expérience 2 : Quel est le meilleur codeur ?

Dans le tableau 4, nous présentons les résultats des trois codeurs présentés dans la section 4 sur les deux ensembles de données. Pour SwDA et MRDA, nous observons qu'une seq2seq équipée d'un codeur hiérarchique surpasse les modèles avec le encodeur RNN, tout en réduisant le nombre de paramètres appris.

Le VGRU$_D$ ne fonctionne pas bien avec l'encodeur PersoHGRU. Lorsqu'il est combiné avec un *mécanisme d'attention guidée*, le PersoHGRU présente une précision compétitive sur SwDA. Cependant, sur MRDA, l'ajout d'une couche de personnalisation nuit à la précision. Cela suggère soit que l'information relative au locuteur n'est pas pertinente pour notre tâche (aucune amélioration observée lors de l'ajout d'une information sur la personne), soit que la hiérarchie considérée n'est pas la structure optimale pour exploiter cette information.

Notre modèle final utilise l'encodeur HGRU car dans la plupart des cas, il présente des performances supérieures.

## 6.3 Expérience 3 : Quel mécanisme d'attention utiliser ?

Le seq2seq encode une phrase source dans un vecteur de longueur fixe à partir duquel un décodeur génère une séquence des AD. L'attention oblige le décodeur à se concentrer davantage sur les phrases sources qui sont pertinentes pour la prédiction d'une étiquette.

Dans NMT [18], le fait de compléter une seq2seq par de l'attention contribue à générer de meilleures phrases. Dans le tableau 4, nous voyons que dans la plupart des cas, l'utilisation d'un simple mécanisme d'attention apporte une amélioration assez faible avec le VGRU et nuit un peu aux performances avec un codeur HGRU. Dans le cas d'une seq2seq composée avec un PersoHGRU et un décodeur sans attention, l'apprentissage échoue : la diminution de la perte d'apprentissage est relativement faible et la seq2seq ne se généralise pas. Il apparaît que dans la classification DA où les séquences sont courtes (5 AD), l'attention n'a pas autant d'impact que dans la classification NMT (qui a des séquences plus longues avec des dépendances globales plus complexes).

Si nous considérons un encodeur HGRU, nous observons que les mécanismes *attention guidée* que nous proposons améliorent la précision du développement, ce qui démontre l'importance de faciliter la tâche en utilisant les connaissances préalables sur l'alignement entre les énoncés et les AD. En effet, lors du décodage, il y a une correspondance directe entre les étiquettes et les énoncés, ce qui signifie que $y_i$ est associé à $u_i$. L'attention guidée douce se concentrera principalement sur l'énoncé actuel avec une petite attention supplémentaire sur le contexte où l'attention guidée dure ne prendra en compte que l'énoncé actuel. L'amélioration due à *attention guidée* démontre que l'alignement entre l'entrée/sortie est une clé avant d'être inclue dans notre modèle.

**Analyse de l'attention:** La figure 2 montre un exemple représentatif des poids d'attention des trois différents mécanismes. La seq2seq avec un mécanisme d'attention normal est caractérisée par une matrice de poids éloignée de l'identité (surtout la partie inférieure droite). Lors du décodage des derniers AD, ce manque d'attention conduit à un AD mal prédite pour un énoncé simple : "Uh-Huh" (*backchannel*). Les deux mécanismes *attention guidée* se concentrent davantage sur la phrase associée a l'AD, à chaque pas de temps, et prédisent avec succès le dernier AD.

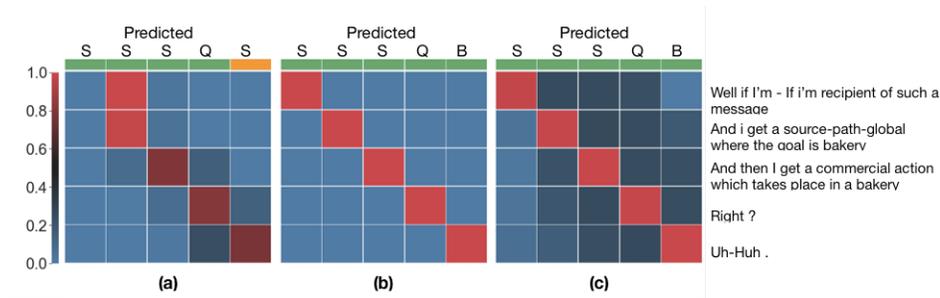

Figure 2: Attention matrix visualisation on MRDA for the fixed context of 5 uttings. La couleur verte pour l'étiquette prédite indique une étiquette correcte, la couleur orange indique une erreur. (a) représente l'HGRU avec attention, (b) représente l'HGRU avec l'attention fortement guidée, (c) est l'HGRU avec l'attention doucement guidée.

Comme le décodeur *attention fortement guidée* présente globalement les meilleurs résultats (à la fois sur SwDA et MRDA) et ne nécessite aucun paramètre supplémentaire, nous l'utiliserons pour notre modèle final.

### 6.4 Expérience 4 : Comment tirer parti de la recherche de faisceaux pour améliorer les performances ?

La recherche de faisceau permet au modèle seq2seq d'envisager des chemins alternatifs dans la phase de décodage.

**Recherche de faisceau pendant la production:** L'utilisation de la recherche par faisceau offre une faible amélioration (amélioration absolue maximale de 0,2%) [3].

Par rapport au NMT, la taille de la sortie est considérablement plus petite ($\mathcal{Y}_{SwDA} = 42$ tandis que $\mathcal{Y}_{MRDA} = 5$) pour la classification AD. Lorsque l'on considère des chemins alternatifs avec un espace de sortie réduit dans des ensembles de données déséquilibrés, la recherche par faisceaux est plus susceptible de considérer des séquences très improbables comme alternatives (par exemple, "s s s s s").

**Entraînement fin:** Comme mentionné précédemment, l'utilisation de la recherche de faisceau pendant l'inférence n'entraîne qu'une amélioration limitée de la précision. Nous réglons finement une seq2seq composée avec un encodeur HGRU et un décodeur avec *attention fortement guidée* (ce modèle a été sélectionné lors des étapes précédentes) avec la perte de niveau de séquence introduite décrite dans la section 4.3. Le tableau 5 montre que cette étape de réglage fin améliore les performances de 1% sur SwDA (84% contre 85%) et de 1,2% sur RMDA (90,4% contre 91,6%).

**seq2seq_BEST:** Notre modèle *seq2seq_BEST* est composé d'un encodeur HGRU et d'un décodeur avec *attention fortement guidée* réglé avec $B_{train} = 2$ et $B_{inf} = 5$ pour SwDA et $B_{train} = 5$ et $B_{inf} = 1$ pour SwDA.

### 6.5 Expérience 5 : Comparaison avec des modèles de pointe

Dans cette section, nous comparons les performances de *seq2seq_BEST* avec d'autres modèles de pointe et analysons les performances des modèles. Le tableau 6 montre les performances du modèle le plus performant seq2seq_BEST sur l'ensemble de test. Le modèle seq2seq_BEST atteint une précision de 85% sur les corpus SwDA. Ce modèle surpasse les modèles Chen et al. [4] et Raheja and Tetreault [21] qui atteignent une précision de 82,9%. En ce qui concerne la MRDA, notre modèle le plus performant atteint une précision de 91,6%, alors que les systèmes actuels de pointe, Chen et al. [4], Kumar et al. [12], atteignent respectivement 92,2% et 91,7%.

---

[3] Les dimensions du faisceau considéré sont petites par rapport aux autres applications [14]. En augmentant la taille du faisceau, on constate que la recherche de faisceau devient très conservatrice [6] et tend à produire des étiquettes fortement représentées dans l'ensemble de formation (par exemple, sd pour SwDA).

|  | SwDA | | RMDA | |
|---|---|---|---|---|
| $B_{inf}$ \ $B_{train}$ | 2 | 5 | 2 | 5 |
| 1 | 84.8 | 84.7 | 91.3 | **91.6** |
| 2 | 84.9 | 84.8 | 91.3 | 91.6 |
| 5 | **85.0** | 84.9 | 91.5 | 91.6 |

Table 5: Précision sur l'ensemble de développement du modèle seq2seq formé avec perte de niveau de séquence. Pour SwDA, le test de Wilcoxon (10 essais avec des graines différentes) a été effectué pour $B_{train} = 2$ et $B_{inf} = 2$ par rapport à tous les autres modèles. Pour le RMDA, le test de Wilcoxon a été effectué (10 essais avec des semences différentes) pour $B_{train} = 5$ et $B_{inf} = 1$ par rapport à tous les modèles avec $B_{train} = 2$.

| Models | SwDa | MRDA |
|---|---|---|
| Li et al. [15] | 82.9 | 92.2 |
| Chen et al. [4] | 81.3 | 91.7 |
| Kumar et al. [12] | 79.2 | 90.9 |
| Raheja and Tetreault [21] | 82.9 | 91.1 |
| seq2seq_BEST | 85.0 | 91.6 |

Table 6: Précision de nos meilleurs modèles (seq2seq) et Baseline_CRF sur les jeux de tests SwDA et MRDA.

## 7 CONCLUSION

Dans ce travail, nous avons présenté une approche nouvelle du problème de la classification des AD. Nous avons montré que notre modèle seq2seq, utilisant un mécanisme *d'attention guidée* nouvellement conçu, obtient des résultats de pointe en remerciant sa capacité à mieux modéliser les dépendances globales.

## ACKNOWLEDGEMENT

Cette recherche a été financée par une bourse de l'Agence Nationale de Recherche Française (ANR-17-MAOI).